\documentclass[conference]{IEEEtran}
\IEEEoverridecommandlockouts
\usepackage{cite}
\usepackage{amsmath,amssymb,amsfonts}
\usepackage{algorithmic}
\usepackage{graphicx}
\usepackage{textcomp}
\usepackage[dvipsnames]{xcolor}

\def\BibTeX{{\rm B\kern-.05em{\sc i\kern-.025em b}\kern-.08em
    T\kern-.1667em\lower.7ex\hbox{E}\kern-.125emX}}
\begin{document}

\title{Robotically adjustable kinematics in a wrist-driven orthosis eases grasping across tasks 



\thanks{
This work was funded by the National Science Foundation via the Graduate Research Fellowship Program (Grant No. DGE 2146752), Trainee Fellowship (Grant No. DGE 2124913), and CAREER (Grant No. 2237843), as well as the Dolores Zohrab Liebmann Fellowship.}
}


\author{\IEEEauthorblockN{Erin Y. Chang, Andrew I. W. McPherson, and Hannah S. Stuart}
\IEEEauthorblockA{Department of Mechanical Engineering, University of California, Berkeley, Berkeley, USA \\ Email: erin.chang@berkeley.edu, drewmcpherson25@berkeley.edu, hstuart@berkeley.edu}}

\makeatletter
\def\ps@IEEEtitlepagestyle{
  \def\@oddfoot{\mycopyrightnotice}
  \def\@evenfoot{}
}
\def\mycopyrightnotice{
  {\footnotesize
  \begin{minipage}{\textwidth}
  \centering
  Copyright~\copyright~2024 IEEE. Personal use of this material is permitted.  Permission from IEEE must be obtained for all other uses, in any current or future media, including reprinting/republishing this material for advertising or promotional purposes, creating new collective works, for resale or redistribution to servers or lists, or reuse of any copyrighted component of this work in other works.
  \end{minipage}
  }
}

\maketitle

\begin{abstract}
Without finger function, people with C5-7 spinal cord injury (SCI) regularly utilize wrist extension to passively close the fingers and thumb together for grasping. 
Wearable assistive grasping devices often focus on this familiar wrist-driven technique to provide additional support and amplify grasp force. 
Despite recent research advances in modernizing these tools, people with SCI often abandon such wearable assistive devices in the long term. 
We suspect that the wrist constraints imposed by such devices generate undesirable reach and grasp kinematics.
Here we show that using continuous robotic motor assistance to give users more adaptability in their wrist posture prior to wrist-driven grasping
reduces task difficulty and perceived exertion. 
Our results demonstrate that more free wrist mobility allows users to select comfortable and natural postures depending on task needs, which improves the versatility of the assistive grasping device for easier use across different hand poses in the arm's workspace. This behavior holds the potential to improve ease of use and desirability of future device designs through new modes of combining both body-power and robotic automation.
\end{abstract}


\section{Introduction}
Dextrous prehensile tasks like grasping objects during activities of daily living (ADLs) can be challenging for people with C5-7 spinal cord injury (SCI). These levels of injury result in tetraplegia and individuals lose function of the hand and fingers, but retain the ability to extend their wrist \cite{Mateo2015}. For some tasks needing only light grip forces, individuals commonly employ the tenodesis grasp technique, where extension of the wrist gently pulls the fingers and thumb toward each other. Assistive devices aimed at improving grasp function often harness this biomechanical coupling strategy for those with residual wrist extension function \cite{Jaeco}. 

These type of passive devices, called Wrist-Driven Orthoses (WDOs), have garnered recent research interest due to the decline in use since their introduction in the 1950s-60s \cite{Nichols1978,Phillips1993}. Recent works presented new designs using accessible materials \cite{Portnova2018, Yeh2023}, and our own prior work showed that adding a motor that reduces the total range of motion (ROM) used in wrist-driven grasping is beneficial and desired by users \cite{Chang2023}. Nevertheless, one component of the early WDO design has yet to be carried over: the actuating lever and locking block that allows users to manually change the setpoint of the wrist angle as needed \cite{Jaeco,Hsu2008}. In past implementation and prior to grasping, users press the actuating lever to decouple the wrist and hand, then move the wrist to one of five discrete positions on the locking block to change the wrist angle. Releasing the actuating lever fixes the setpoint of the wrist at their desired position for wrist-driven grasping.

In this work, we investigate an additional role for the motor that allows the user to automatically vary the wrist's setpoint with natural wrist movement. By conveniently automating wrist setpoint adjustment with motorized assistance, we aim to eliminate troublesome manual adjustments and instead enable the user to easily perform a wider variety of unimanual grasps. 
Since the position of the wrist affects the biomechanics of the rest of the upper body \cite{Montagnani2015}, we anticipate that 
comfort and task difficulty will depend on the wrist's posture when the hand approaches the object to grasp. In some instances, one might prefer wrist flexion, while in others, wrist extension may be more effective. Therefore, we hypothesize that the ability to easily choose a wrist posture based on task needs will improve users' grasp function versatility, or the variety of tasks they can comfortably perform. 

\section{Methods}
\begin{figure}[!t]
\centering
\includegraphics[width=3.45in]{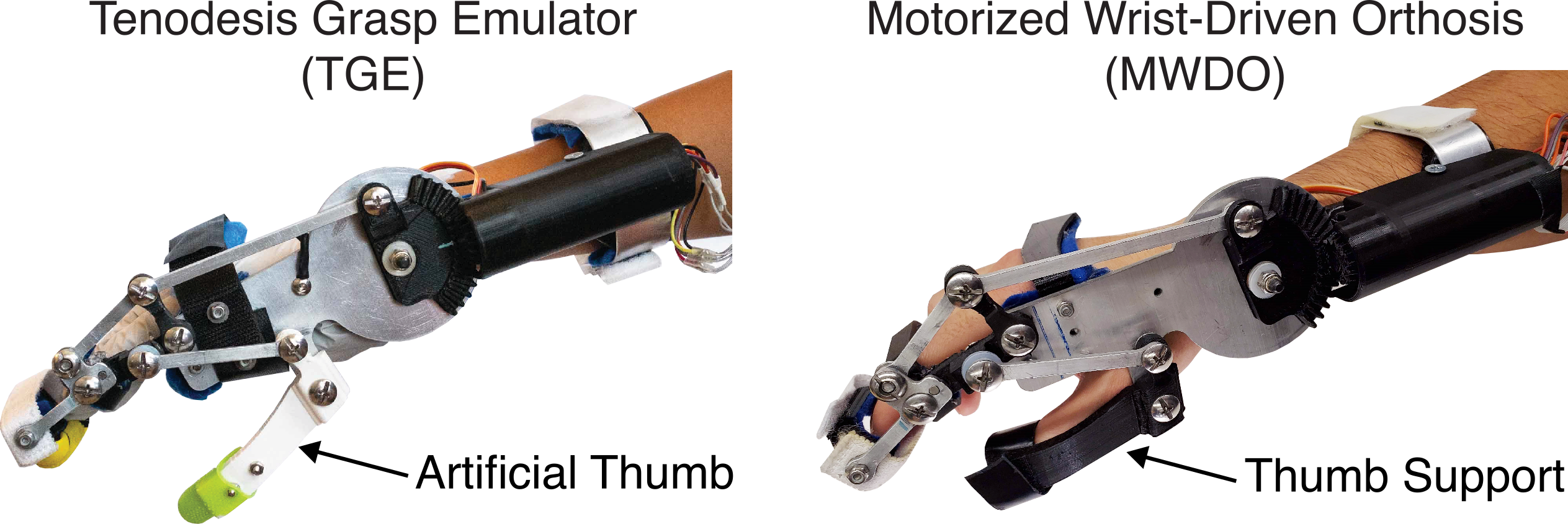}
\caption{Assistive devices used in this study. The Tenodesis Grasp Emulator (TGE) \cite{Chang2022} was worn by subjects with normative hand function and the Motorized Wrist Driven Orthosis (MWDO) \cite{McPherson2020} was worn by the subject with SCI. Images adapted from these prior publications.}
\label{fig:devices}
\vspace{-4mm}
\end{figure}

\begin{figure*}[!t]
\centering
\includegraphics[width=5.8in]{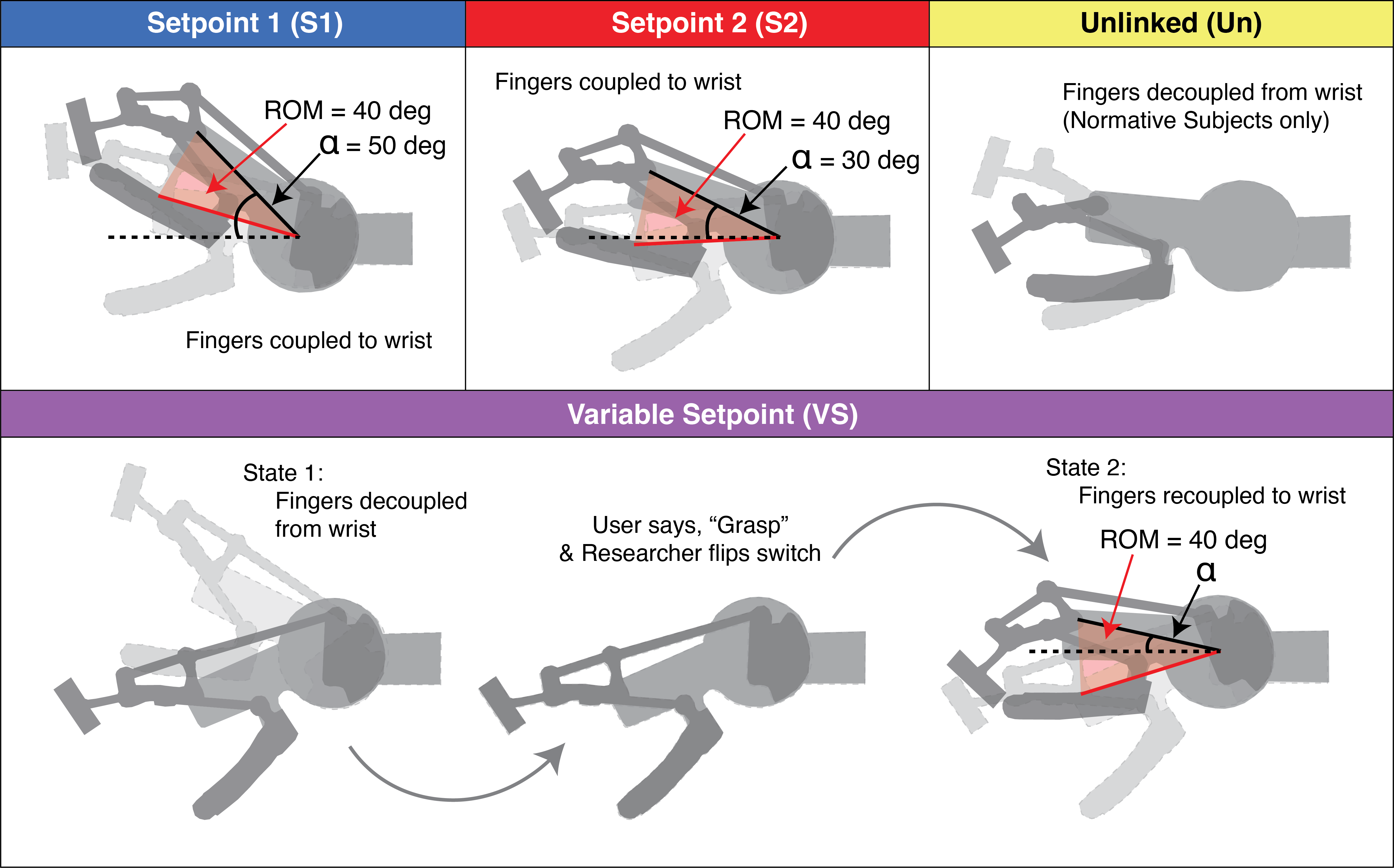}
\caption{Actuation modes implemented into the assistive devices. The motor in S1 mode and S2 mode limits the wrist's ROM to 40 deg during wrist-driven grasping, and the setpoint of the wrist at the closed hand posture ($\alpha$) stays fixed at 50 deg and 30 deg, respectively. One link is disconnected in the Unlinked mode such that grasping is accomplished with finger flexion. The motor in VS mode takes on dual roles: in State 1 it allows the wrist to move freely without affecting hand posture, and in State 2 it assists with wrist-driven grasping by limiting the wrist's ROM to 40 deg. $\alpha$ in VS mode is determined by the wrist posture selected by the user at the end of State 1.}
\label{fig:devicemodes}
\vspace{-4mm}
\end{figure*}

Three subjects with normative hand function and one subject with C5-6 SCI were recruited under the University of California, Berkeley Institutional Review Board approved protocol \#2020-02-12983 to perform a series of grasp and release tasks. Eligible normative subjects had no injuries to their upper limbs in the last six months. Eligible subjects with SCI had been living stably with SCI between C5-7 for the last six months and had wrist extension ability against gravity. All subjects wore an assistive device on their right hand and wrist, described in Sec. \ref{sec:devices} and pictured in Fig. \ref{fig:devices}. Details on the design and technical specifications of each device can be found in prior works \cite{Chang2022, McPherson2020}.

\subsection{Assistive Device}
\label{sec:devices}
We outfitted the testbeds from our prior work with new control modes for this study. Subjects with normative hand function wore the Tenodesis Grasp Emulator (TGE) from \cite{Chang2022}, utilizing an artificial thumb to avoid unintentionally squeezing the thumb and fingers together during wrist-driven prehension, and the subject with SCI wore the Motorized Wrist-Driven Orthosis (MWDO) from \cite{McPherson2020} (Fig. \ref{fig:devices}). 

In these prior works, we previously created a ``shared" grasp system combining passive body-power aspects of a WDO with an active robotic motor component, to harness both the advantages of traditionally passive and traditionally active systems \cite{Chang2023}. Implementing the shared system in the MWDO reduced the total wrist ROM between the open and closed hand postures to 40 deg, compared to 70 deg when the MWDO used the passive system alone. Since users with SCI preferred using the shared system, we continue to implement it in the present work. In the preceding studies, we ``fixed" the setpoint of the wrist, such that the user's closed hand wrist posture remained constant for all tasks. In contrast, the actuating lever and locking block mechanism in early WDO designs creates a discretely ``variable" setpoint device, where the user can change the setpoint to accommodate various task needs. 

In this work, all subjects experienced three motorized device modes pictured in Fig. \ref{fig:devicemodes}: Setpoint 1 (S1), Setpoint 2 (S2), and Variable Setpoint (VS) mode. In S1 mode, the wrist setpoint in the closed hand posture is fixed at $\alpha$ = 50 deg wrist extension, and in S2 mode, the wrist setpoint is fixed at $\alpha$ = 30 deg wrist extension. S1 and S2 can be likened to two different locking block positions toward the ends of the wrist ROM in the subject with SCI. In the VS mode, the device uses a switch-activated state machine. In State 1, the motor rotates the link perpendicular to the wrist joint in the opposite direction of wrist movement to maintain a constant hand posture while the user can freely move the wrist. In State 2, the motor supports grasping similar to S1 or S2 mode, but at the setpoint last selected in State 1 (Fig. \ref{fig:devicemodes}). To use the VS mode, the user starts in State 1 with the fingers open, and positions their wrist and hand as desired to grasp the object in the first target location. The user then says ``Grasp," and the researcher switches the device to State 2, locking in this implicitly selected setpoint for the user to perform the wrist-driven grasp, transport the object, and release it in the second target location. Prior to the next task, the researcher resets the device to State 1. Unlike in the locking block mechanism, VS mode generates a continuously changing setpoint during the reaching-to-grasp phase.

In addition to the motorized device modes, the subjects with normative hand function experienced the Unlinked (Un) mode (Fig. \ref{fig:devicemodes}). In this mode, the motor is turned off and one linkage pin is disconnected to fully decouple the thumb from the wrist and motor movement as a benchmark for normative (non-wrist-driven) grasping while wearing the device. While wearing the device in the Un mode, users can flex and extend their wrist independently of the fingers. Since the fingers remain coupled to the artificial thumb, flexing and extending the fingers closes and opens the hand for grasping.

\begin{figure}[!t]
\centering
\includegraphics[width=3.3in]{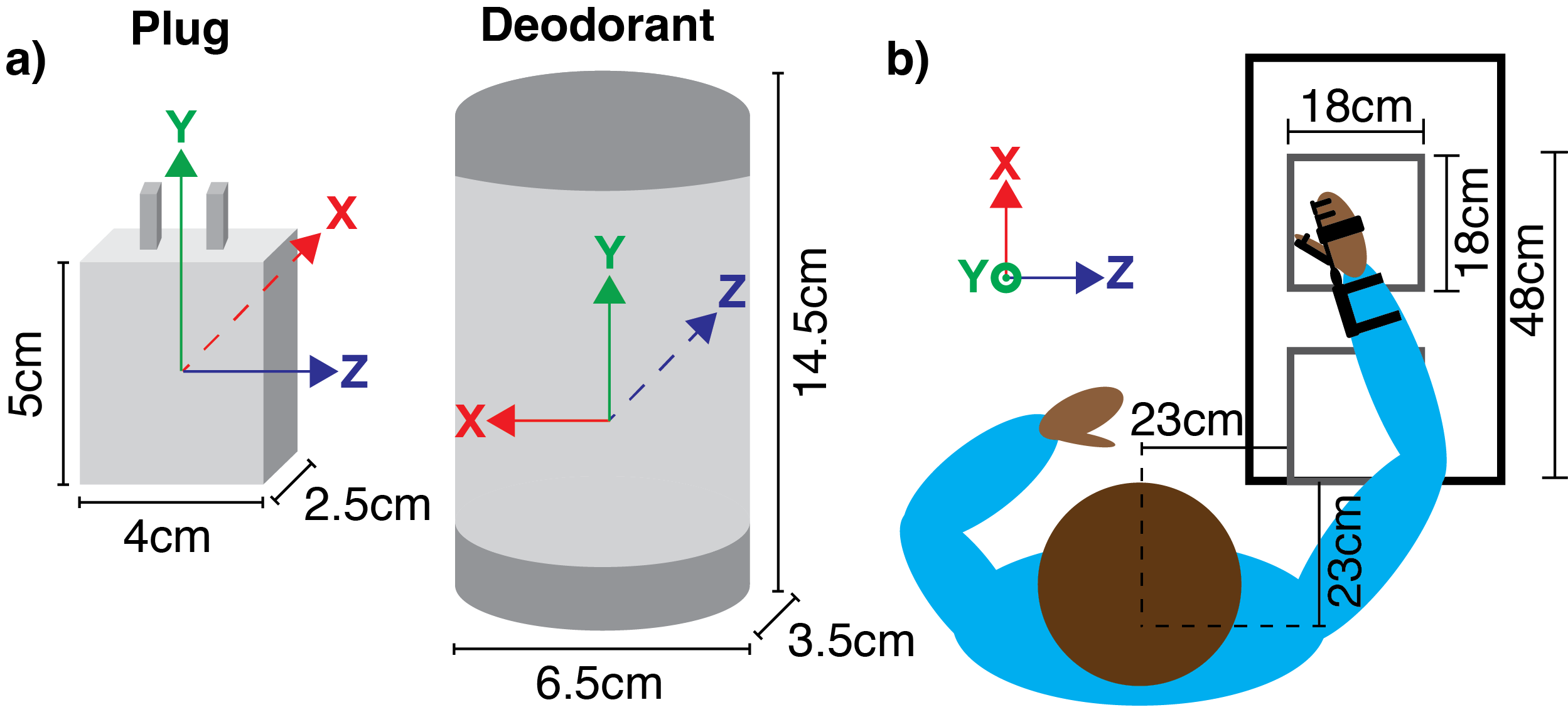}
\caption{Task specifications: a) Subjects grasped a plug over the top with their hand closing around the z axis, and a stick of deodorant, from the side with their hand closing around the y axis. b) Subjects moved objects between two target locations: one \textit{near} the body and one \textit{far} from the body.}
\label{fig:tasks}
\vspace{-4mm}
\end{figure}

\subsection{Experimental Procedure}
\label{sec:procedure}
As shown in Fig. \ref{fig:tasks}a, we selected a rectangular plug and a deodorant stick, two objects that elicit different wrist orientations and hand postures and may be encountered during ADLs. In this experiment, subjects manipulated the plug with the flat side facing down to focus simply on grasping and releasing without requiring additional forces needed to push it into an outlet. We expect that grasping the plug over the top (about the Z axis), such as when plugging a device into a power strip, likely requires wrist angles closer to flexion. In contrast, we expect that grasping a deodorant stick from the side (about the y axis), such as when grabbing it for use from a counter or shelf, likely requires more wrist extension. At the grasp and release workspace shown in Fig. \ref{fig:tasks}b, subjects with normative hand function sat in a fixed height stool and the subject with SCI sat in their personal wheelchair. The table was also maintained at a fixed height for all participants.

Subjects moved each object between the near and far target locations ten times in each device mode (where grasp in location 1 and release in location 2 is considered one trial), resulting in a total of 80 trials for subjects with normative hand function and 60 trials for the subject with SCI. Task failures 
were noted by a researcher, then omitted from remaining analyses. When a task failure occurred, the subject reset the task and repeated it until it was completed successfully.

During these grasping tasks, the Impulse X2E Motion Capture system (PhaseSpace, San Leandro, California, USA) recorded upper body motion and a Hero9 Black camera (GoPro, Inc., San Mateo, California, USA) recorded video of the frontal plane. Following each set of ten trials with one object and one device mode, subjects reported their perceived exertion using the Borg CR-10 scale \cite{borg1998} and described how they felt the device mode affected their upper body movement. After they experienced all the device modes for each object, they additionally ranked the modes in order of difficulty.

\subsection{Data Analysis}
Subject ratings of perceived exertion and difficulty were normalized to each subject's VS mode score, since all subjects experienced this mode and it represents a self-selected setpoint. This method allowed us to visualize how S1, S2, and Un modes compared to VS mode when baseline perceived exertion varies between subjects. Motion capture data was analyzed using the methods described in \cite{Chang2023}.


The Shapiro-Wilk test determined 71\% of data distributions were normal and Levene's test determined 75\% of groups were homoscedastic. Thus, we proceeded to perform statistical analysis using Analysis of Variance (ANOVA). For significant effects, we applied Tukey's honestly significant difference post-hoc test during pairwise comparisons between modes. 

\section{Results}
\begin{figure*}[!t]
\centering
\includegraphics[width=5.45in]{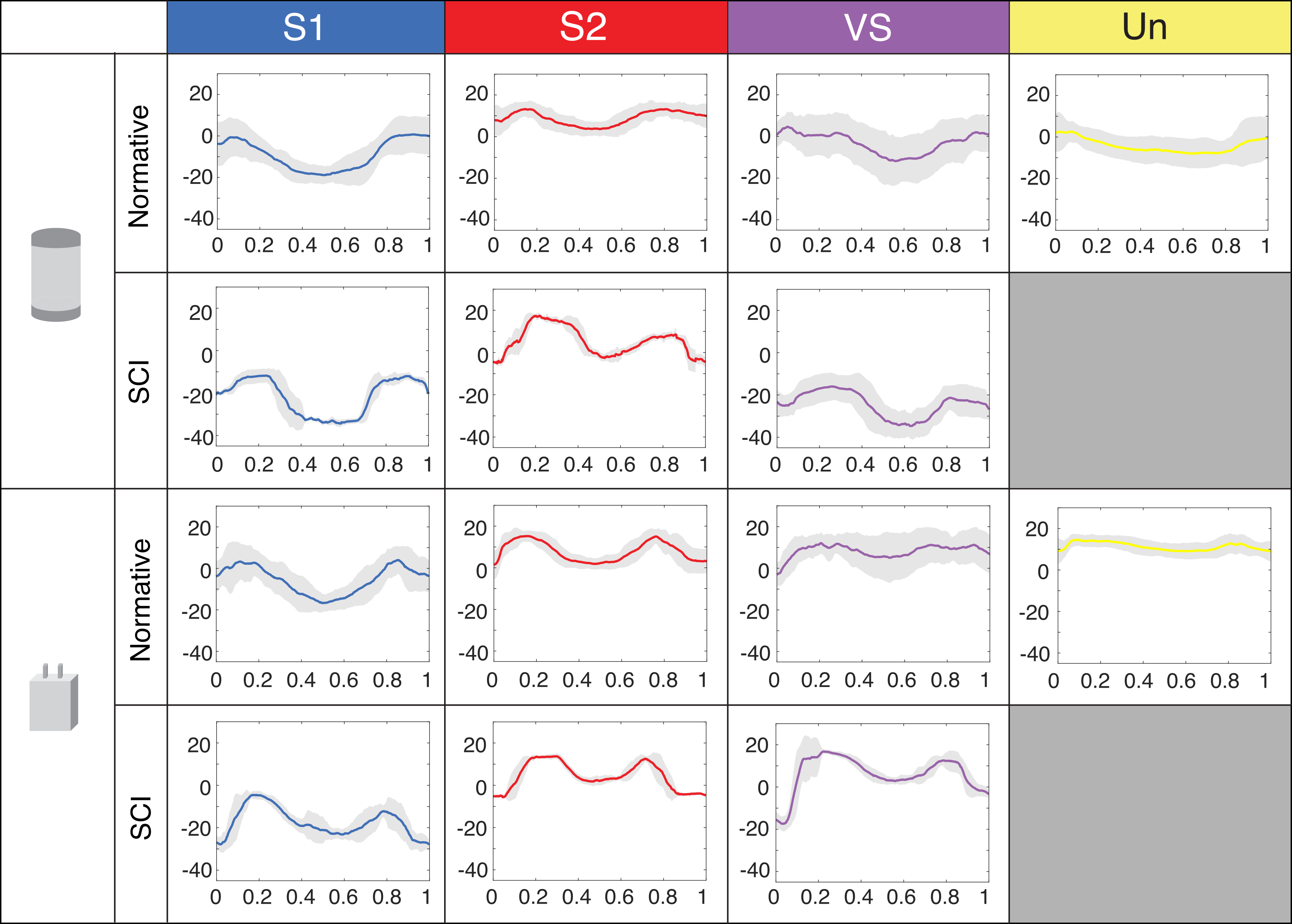 }
\caption{Subjects' mean (color) and standard deviation (shaded) of wrist flexion (+) and extension (-) angles in degrees over normalized task time.}
\label{fig:WFEtraj}
\end{figure*}

\begin{figure*}[!t]
\centering
\includegraphics[width=5.65in]{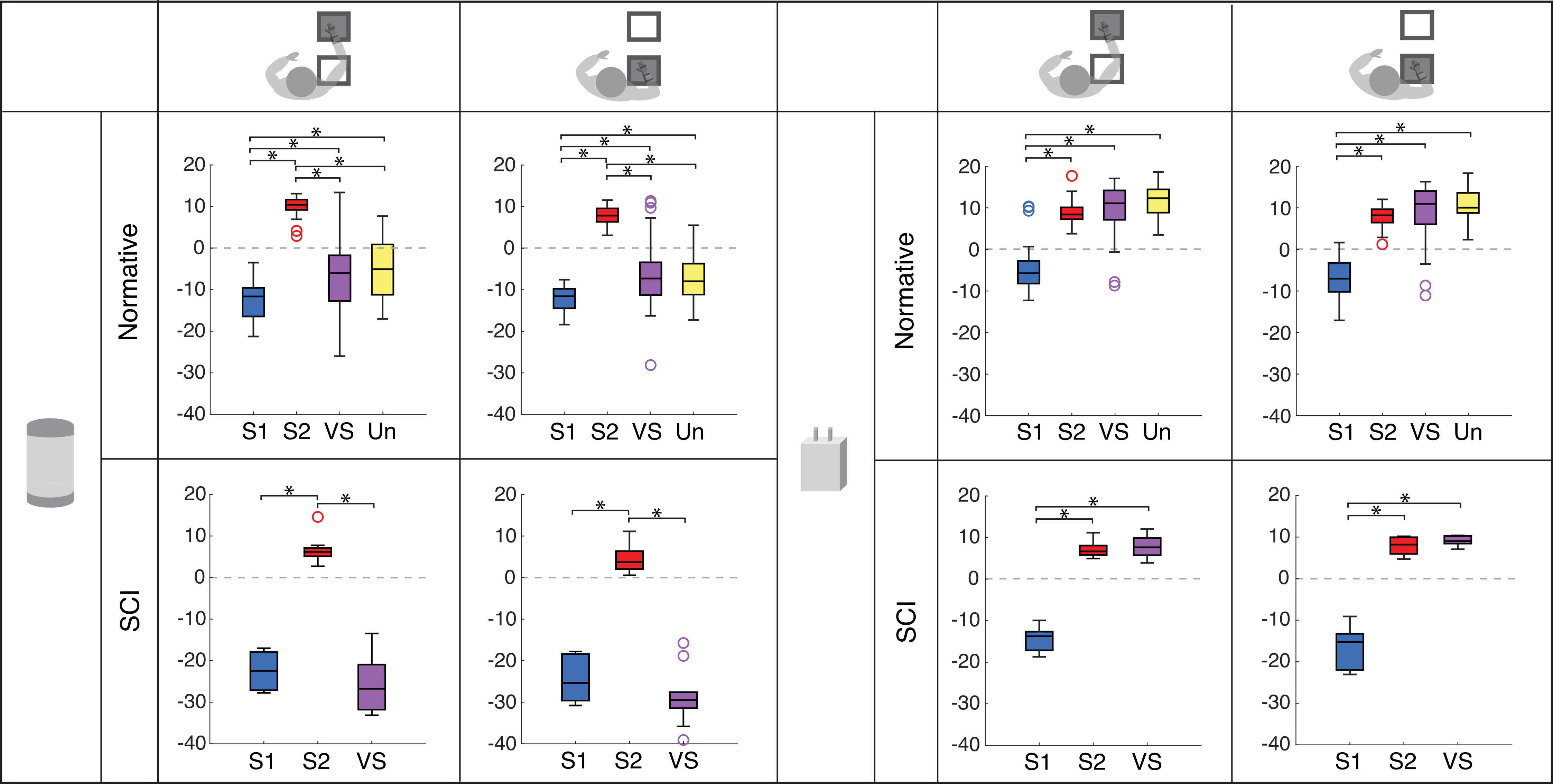 }
\caption{Subjects' wrist flexion (+) and extension (-) angles in degrees, when manipulating each object at each location. (*p$<$0.05, empty circles = outliers)}
\label{fig:WFEangles2}
\vspace{-4mm}
\end{figure*}

\subsection{Wrist Angles}
Fig. \ref{fig:WFEtraj} shows the wrist flexion and extension angles as they change over the course of a grasp and release task, starting and ending with the hand resting on the table. Prior to grasping objects in S1, S2, or VS mode, users often needed to flex their wrist to open their hand, indicated by the first peak in these trajectories. Similar flexion occurred when releasing objects with these modes, represented by the second peak in these trajectories. Peak wrist angles are additionally represented in Fig. \ref{fig:WFEangles2}, showing that subjects exhibit similar wrist angle patterns when manipulating objects in each location. 
 
When grasping the deodorant, normative subjects chose a mean wrist angle close to a neutral posture of 1.74 deg flexion, while the subject with SCI chose a larger mean wrist angle of 16 deg extension. When grasping the plug, normative subjects chose a mean wrist angle of 11.73 deg flexion and the subject with SCI chose a mean wrist angle of 16.79 deg flexion. Naturally, normative subjects using the Un mode likewise placed their wrist close to a neutral posture during the deodorant tasks (mean: 2.06 deg extension) and flexion during the plug tasks (mean: 13.87 deg flexion).

As we predicted with VS mode, subjects selected wrist postures closer to those used with S1 mode for the deodorant task and postures closer to those used with S2 mode for the plug task. Statistical testing indicated that the wrist angles used to grasp the deodorant stick in S2 mode were statistically significantly different from wrist angles when grasping in any other mode. Similarly, wrist angles used to grasp the plug in S1 mode were statistically significantly different from wrist angles when grasping in any other mode.

\begin{figure*}[!t]
\centering
\includegraphics[width=5.5in]{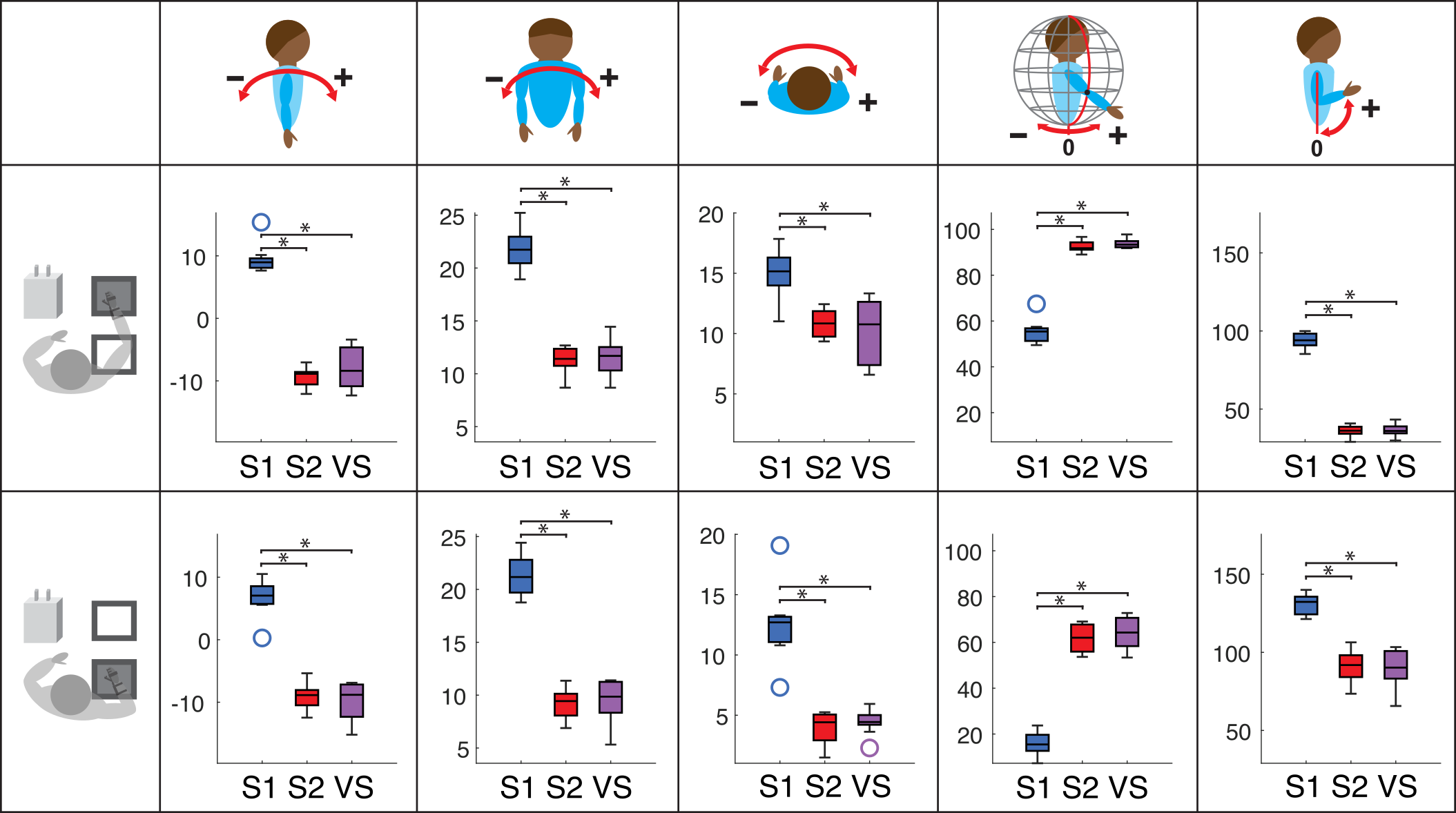 }
\caption{Upper body angles in degrees of the subject with SCI, at grasp and release instances for the plug in the each location. From left to right: trunk backward extension/forward flexion, trunk lateral flexion, trunk rotation, glenohumeral plane of elevation, elbow flexion. (*p$<$0.05, empty circles = outliers)}
\label{fig:s2angles}
\vspace{-4mm}
\end{figure*}

\begin{figure}[!t]
\centering
\includegraphics[width=3.3in]{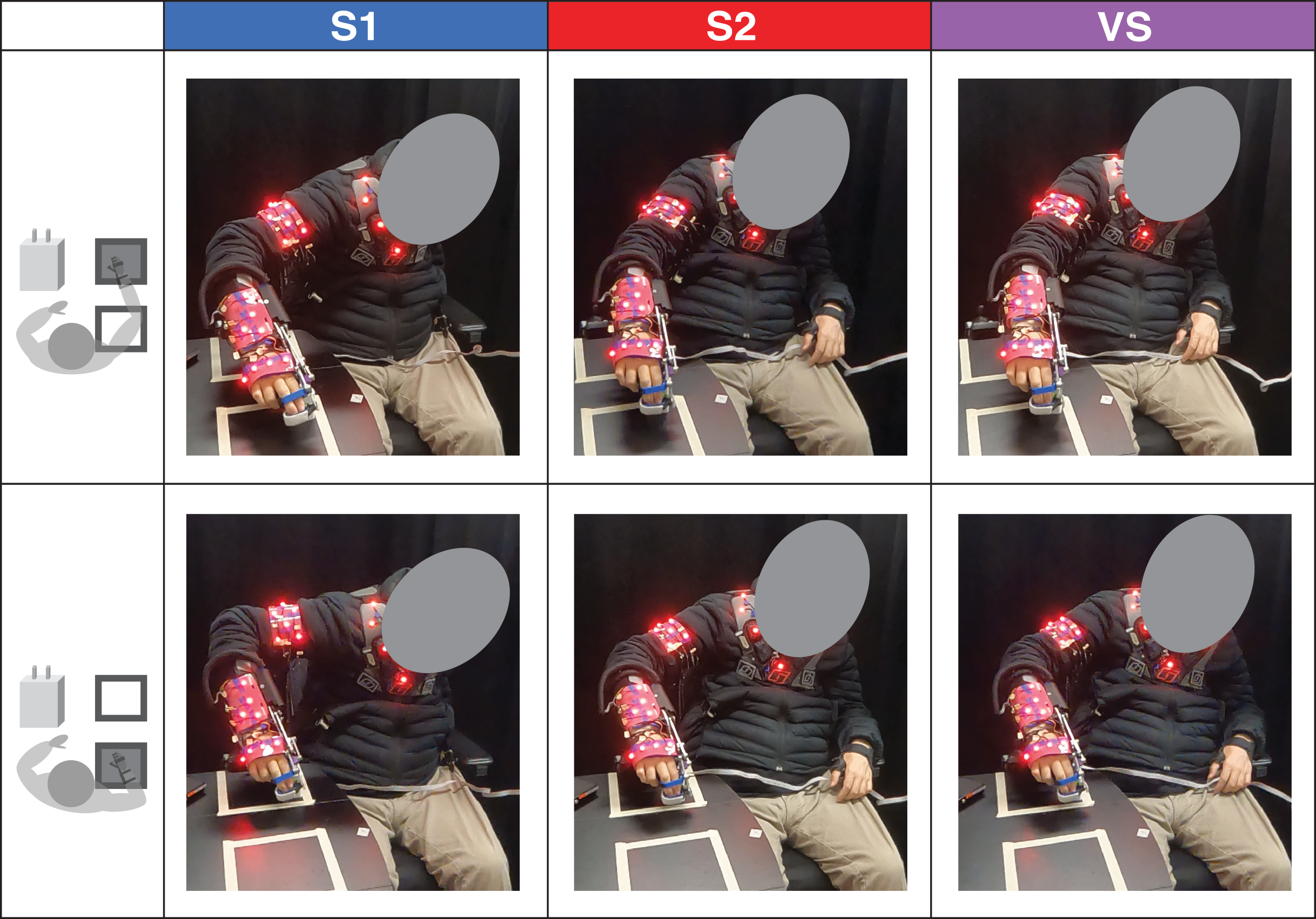 }
\caption{Postures of the subject with SCI, when grasping the plug in each location. Compensation trends appear similar between S2 and VS mode.}
\label{fig:postures}
\vspace{-6mm}
\end{figure}

\subsection{Upper Body Compensation}
The subject with SCI exhibited statistically significant patterns of upper body compensation that varied between modes during the instances of grasping and releasing the plug in each location, shown in Fig. \ref{fig:s2angles}. When manipulating the plug, S2 and VS mode showed similar trends in compensation. Compared to S1 mode, these two modes resulted in trunk extension against backrest of the chair instead of forward trunk flexion, and reduced lateral trunk flexion, trunk rotation, and elbow flexion, but increased glenohumeral plane of elevation. Examples of these postures are shown in Fig. \ref{fig:postures}.

When manipulating the deodorant, the subject with SCI showed similar trends between S1 and VS mode, but with smaller differences in angular magnitude. Trunk angles often differed by 10 deg or less and the glenohumeral plane of elevation and elbow flexion angles differed by 20 deg or less. Subjects with normative hand function did not exhibit distinct changes in upper body movement between the modes.

\begin{figure*}[!t]
\centering
\includegraphics[width=5.4in]{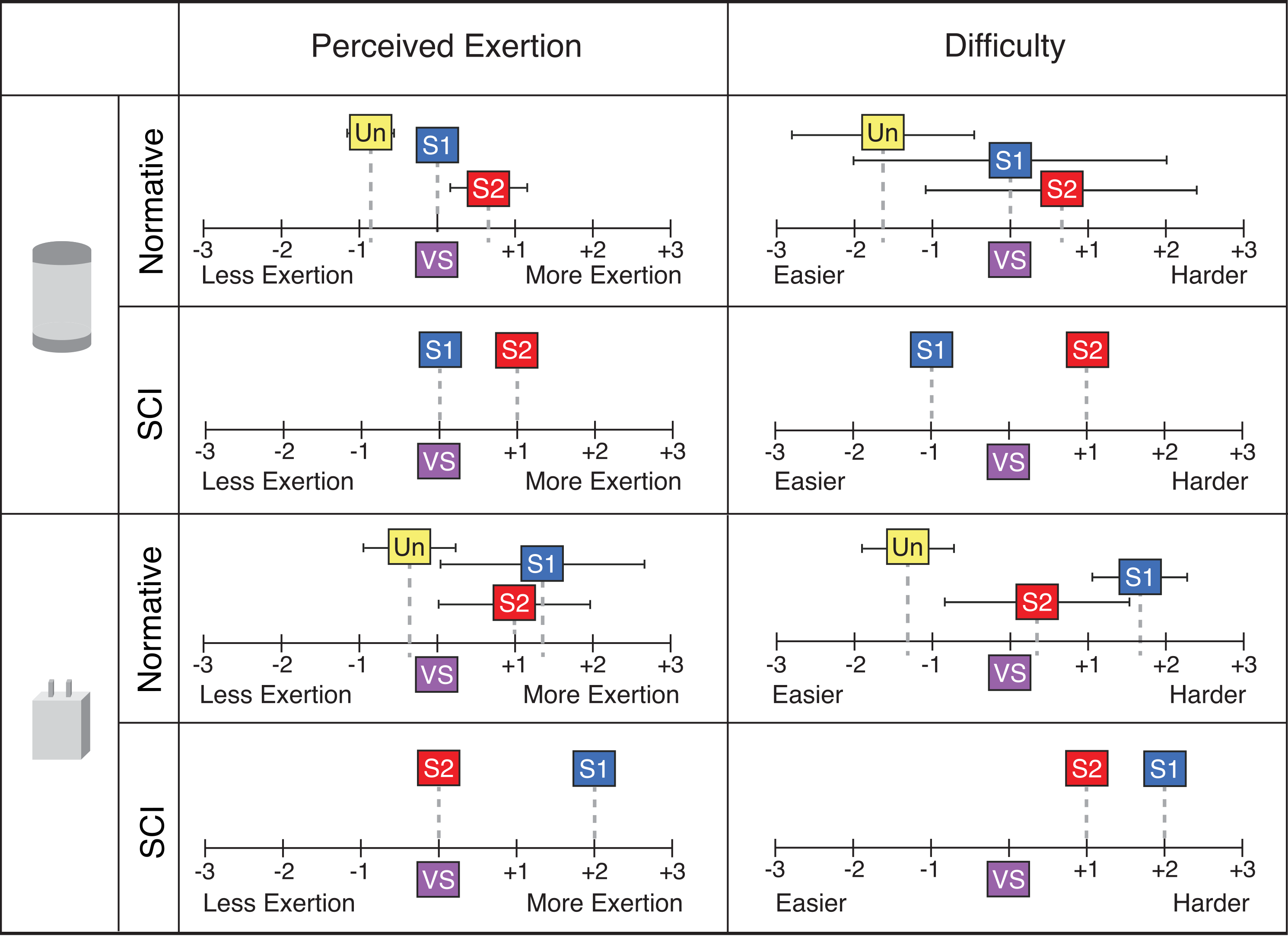}
\caption{Subjects' perception of each device mode for each object. Column 1 shows the relative change (average and standard deviation) in Borg CR-10 score for each device mode, normalized to the VS mode score. Column 2 shows the relative change (average and standard deviation) in difficulty rating for each device mode, normalized to the difficulty of VS mode.}
\label{fig:Preferences}
\vspace{-4mm}
\end{figure*}

\subsection{Ratings of Exertion and Difficulty}
Each fixed setpoint mode resulted in notably more difficulty and exertion for one task than the other, which aligned with our initial predictions. On average, subjects perceived the most exertion with S2 mode when manipulating the deodorant and S1 mode when manipulating the plug (Column 1 of Fig. \ref{fig:Preferences}). Similarly, they rated S2 as the most difficult mode for the deodorant and S1 as the most difficult mode for the plug (Column 2 of Fig. \ref{fig:Preferences}). As expected, normative subjects perceived the least exertion when they operated the baseline Unlinked mode and similarly rated this mode the easiest.

Subjects tended to perceive the same or less exertion with the VS mode compared to the other motorized modes. During the deodorant tasks, subjects perceived the same amount of exertion with the VS and S1 modes, both of which were less than the exertion perceived with S2 mode. During the plug tasks, they perceived less exertion with the VS mode than either of the fixed setpoint modes, with the exception of the subject with SCI, who percieved similar exertion between VS and S1 mode. 
Likewise, subjects tended to rate the VS mode as easier or of similar difficulty to the other motorized modes. During the deodorant tasks, VS mode often ranked between the S1 and S2 modes. 
During the plug tasks, VS mode often ranked easier than the two fixed setpoint modes.
Consistent with our hypothesis, these results indicate that users found grasping with the VS mode to be easier and require less exertion than either of the fixed setpoint modes when considering performance across both tasks.

\section{Discussion}
When choosing a wrist angle during reach-to-grasp in the VS mode, subjects typically selected neutral or extension angles for the deodorant and flexion angles for the plug, as we hypothesized. Similarly, during the deodorant task, starting with a large wrist extension angle was perceived as easier and requiring less exertion, and during the plug tasks, less wrist extension was perceived as easier and requiring less exertion.


Subjects indicated that constraining the wrist closer to flexion during the deodorant task and extension during the plug task was challenging and undesirable. With the deodorant, subjects expressed frustration with limited wrist extension in S2 mode, often citing that their arm and torso collided when attempting to position the hand, and that it took more effort to open the hand with these wrist postures. Similarly, with the plug, subjects commonly shared that excess wrist extension in the S1 mode required them to largely move their torso in order to position the hand, which they strongly disliked.

These findings suggest that despite the desirable addition of robotic assistance to reduce wrist ROM
\cite{Chang2023}, fixed setpoints like those implemented in the S1 and S2 mode make it difficult to comfortably perform a wide variety of tasks. 
Though our initial approach toward a continuously variable setpoint required researcher assistance to switch between motorized states, subjects with and without SCI found it easier and more comfortable to accomplish the set of both tasks with the VS mode. In contrast, the suitability of a fixed setpoint depends on the context; it can be advantageous in one situation, but poses drawbacks in another.

While such initial findings indicate that free control over wrist posture when using a wrist-driven device is favorable, future research should investigate the impact of a continuously variable setpoint control system (like VS mode) in unstructured environments like participants' homes. Generalizing these preliminary results to the broader population of individuals with SCI should be performed with caution, since limited subjects participated in this study and we performed testing in a short-term, structured lab environment.

\section{Conclusion}
We demonstrated a promising new modality of robotic assistance in a wrist-driven device by enabling users to freely and continuously orient their wrist prior to grasping. With just two household objects, we showed that continuous robotically moderated setpoint control enabled users to accomplish more activities with improved comfort and reduced difficulty than a single fixed setpoint alone. Preliminary findings showed further promise in reducing upper body compensation and awkward postures in individuals with SCI, which can lead to musculoskeletal pain and injuries.

With advances in robotic automation and the user in mind, future work should investigate alternative implementations for switching between the two states that comprise the VS mode. While user-accessible buttons or switches may be a feasible solution, exploring new intuitive methods that reduce cognitive and manual effort may 
impact long-term device adoption.

\section*{Acknowledgment}
Any opinions, findings, and conclusions or recommendations expressed in this material are those of the authors and do not necessarily reflect the views of the National Science Foundation. The authors acknowledge the support of the members of the Embodied Dexterity Group.

\bibliographystyle{IEEEtran}
\bibliography{bibliography}

\end{document}